\documentclass[preprint,review,11pt,authoryear]{elsarticle}

\usepackage{subcaption}
\usepackage{graphicx}
\usepackage{lipsum}

\usepackage{amssymb}
\usepackage{amsthm}

\usepackage{amsmath}
\usepackage{eurosym}

\newcommand{\xb}{\mathbf{x}}
\newcommand{\Xb}{\boldsymbol{X}}
\makeatletter
\newcommand\bigLozenge{\mathop{\mathpalette\bigL@zenge\relax}}
\newcommand\bigL@zenge[2]{%
	\vcenter{\hbox{\m@th
			\scalebox{\ifx#1\displaystyle 2\else1.2\fi}{$#1\blacklozenge$}%
	}}%
}

\usepackage[left=2.8cm,right=2.8cm,top=2.8cm,bottom=2.8cm]{geometry}

\def\ps@pprintTitle{%
	\let\@oddhead\@empty
	\let\@evenhead\@empty
	\def\@oddfoot{}%
	\let\@evenfoot\@oddfoot}
\makeatother

\begin{document}

\begin{frontmatter}



\title{Profit Driven Decision Trees for Churn Prediction}


\author[MathematicsAddress]{Sebastiaan H\"oppner}
\ead{sebastiaan.hoppner@kuleuven.be}
\author[EconomicsAddress]{Eugen Stripling}
\ead{eugen.stripling@kuleuven.be}
\author[EconomicsAddress,SouthamptonAddress]{Bart Baesens}
\ead{bart.baesens@kuleuven.be}
\author[EconomicsAddress]{Seppe vanden Broucke}
\ead{seppe.vandenbroucke@kuleuven.be}
\author[MathematicsAddress]{Tim Verdonck}
\ead{tim.verdonck@kuleuven.be}

\address[MathematicsAddress]{KU Leuven, Department of Mathematics, Celestijnenlaan 200B, 3001 Leuven, Belgium}
\address[EconomicsAddress]{KU Leuven, Faculty of Economics and Business, Naamsestraat 69, 3000 Leuven, Belgium}
\address[SouthamptonAddress]{University of Southampton, School of Management, Highfield Southampton, SO17 1BJ, United Kingdom}

\begin{abstract}
Customer retention campaigns increasingly rely on predictive models to detect potential churners in a vast customer base. From the perspective of machine learning, the task of predicting customer churn can be presented as a binary classification problem. Using data on historic behavior, classification algorithms are built with the purpose of accurately predicting the probability of a customer defecting. The predictive churn models are then commonly selected based on accuracy related performance measures such as the area under the ROC curve (AUC). However, these models are often not well aligned with the core business requirement of profit maximization, in the sense that, the models fail to take into account not only misclassification costs, but also the benefits originating from a correct classification. Therefore, the aim is to construct churn prediction models that are profitable and preferably interpretable too. The recently developed expected maximum profit measure for customer churn (EMPC) has been proposed in order to select the most profitable churn model. We present a new classifier that integrates the EMPC metric directly into the model construction. Our technique, called ProfTree, uses an evolutionary algorithm for learning profit driven decision trees. In a benchmark study with real-life data sets from various telecommunication service providers, we show that ProfTree achieves significant profit improvements compared to classic accuracy driven tree-based methods.
\end{abstract}


\end{frontmatter}


\section{Introduction}\label{Introduction}
Companies operating in saturated markets are continuously challenged to retain their customers. Besides spending resources on attracting new customers, they try to prevent existing customers from defecting. To that end, customer retention campaigns aim to identify which customers intend to switch to a competitor and present them with an incentive offer to remain with the company. However, detecting potential churners out of typically millions of customers is a difficult task. For that reason, companies increasingly rely on predictive churn models to remain competitive. Churn models aim to predict a customer's churning propensity, also called churn score, by using behavioral and historical information. Yet, these models often focus on achieving maximum prediction accuracy rather than aiming their attention at the most important business requirement: profit maximization. For a retention campaign to be successful, it is not only crucial to correctly identify would-be churners, but also to detect those who are the most profitable to the business and thus worth retaining. The ideal churn prediction model is therefore capable of effectively identifying churners and simultaneously taking profit concerns of the business into account.

Traditionally, the performance of a churn model is evaluated using accuracy related measures, which do not take profit maximization into concern. For instance, for binary classification problems, a popular choice for selecting the winning model is the area under the ROC curve (AUC), because of its simplicity and intuitive interpretation. However, it has been shown by \citet{hand2009measuring} that the AUC implicitly assumes that misclassification errors carry the same costs which also alter across classifiers. Reasonably, the cost of misclassifying a non-churner as a churner is different from the cost incurred when an actual churner is classified as a non-churner. Furthermore, misclassification costs are  ultimately a property of the classification problem, and should be independent of the applied classifier. Besides taking costs into concern, it is also generally recommended to incorporate the benefits of making a correct classification into the performance metric \citep{elkan2001foundations}.

For predictive churn models, \citet{verbraken2013novel} proposed a profit-based performance metric, called the \textit{expected maximum profit measure for customer churn} (EMPC), which allows identifying the most profitable model. Performance is measured based on the average classification profit in which the specified costs and benefits are the ones associated with a retention campaign. Additionally, the proposed cost benefit framework provides the \textit{expected profit maximizing fraction for customer churn} ($\overline{\eta}_{empc}$) that determines the optimal fraction of the customer base to target in the retention campaign. \citet{verbraken2013novel} showed that there are great discrepancies between the EMPC and AUC when assessing a model's performance, and that the use of AUC as a model selection criterion leads to suboptimal profits. For customer retention campaigns, it is therefore recommended to select the winning model based on the EMPC in order to achieve maximum profit. 

Although the EMPC enables a profit-based model evaluation, profit concerns are not directly integrated into the model construction. Therefore, we propose a new profit maximizing classifier, called ProfTree, that optimizes the EMPC in its training step. In our approach, a classification tree method is utilized to estimate churn scores, that is, the probability of a customer leaving the company. These scores are required for computing the profit measure. We opt to use decision trees because they are typically easy to interpret and fast to compute. This is especially useful in a churn prediction setting to understand why customers defect and work out corresponding churn prevention strategies. The split rules of the decision tree are optimized according to the EMPC metric using an evolutionary algorithm. The choice for the usage of a global optimization method like an evolutionary algorithm is justified because classical forward-search recursive partitioning methods, such as CART \citep{breiman1984classification} and C4.5 \citep{quinlan1993c45}, will only yield locally optimal solutions. 

The remainder of this paper is organized as follows. In Section \ref{sec:preliminaries}, we discuss the essential building blocks that are relevant to ProfTree as well as related work. Section \ref{sec:ProfTree} outlines the detailed explanations of our research contributions. Section \ref{sec:casestudy} illustrates the usage of ProfTree on a real-life customer churn data set and compares our method with other well-known classifiers. Section \ref{sec:empiricalevaluation} describes the experimental setup and the results of an extensive benchmarking study as well as a discussion thereof. Finally, we give some concluding remarks and potential directions for future research in Section \ref{sec:conclusion}.

\section{Preliminaries\label{sec:preliminaries}}\label{Preliminaries}

\subsection{Customer Churn Prediction Formulated as a Classification Problem\label{subsec:classificationproblem}}

The task of predicting customer churn can be formulated as a binary classification problem. The aim is to assign instances (i.e. customers) to one of the two classes $Y=\{\text{no churn, churn}\}$ on the basis of their observed features $\xb\in\Xb$. Some examples of features that are used in retention campaigns are the lenght of the customer's contract period, the average revenue, the number of filed complaints and the region where the customer lives. The objective of classifiers is to discriminate instances from both classes as accurately as possible.  A popular method for dealing with binary classification problems is a decision tree because it is easy-to-use and offers high interpretability. Furthermore, classification trees are able to cope with complex data structures such as nonlinearities and don't need dummy coding or complex transformations to handle categorical features.

A classification tree aims at modeling the binary response variable $Y$ by a vector of $p$ predictor variables $\Xb=\left(X_1,\ldots,X_p\right)$.
Throughout the text, we encode ``churn'' as $1$ and ``no churn'' as $0$, so $\{\text{no churn, churn}\}$ becomes $\{0,1\}$. Tree-based methods first partition the feature space $\mathbf{X}$ into a set of $M$ rectangular regions $R_m$ ($m=1,\ldots,M$) based on split rules, and then fit a (typically simple) model within each region $\{Y|\mathbf{X}\in R_m\}$, e.g. a constant like the mode. In this section, we will closely follow the notation used by \citet{grubinger2011evtree} to describe the partitioning algorithm, the parameter space and the optimization problem. As done by \citet{grubinger2011evtree}, we only consider tree models with two-way splits and with some maximum number of terminal nodes $M_{max}$. This does not really restrict the problem since multiway splits are equivalent to a sequence of two-way splits and the maximal size of the tree is always limited by the number of instances in the training sample. Throughout the text, we denote a classification tree with $M$ terminal nodes by
\begin{equation}\label{tree}
\theta=\left(v_1,s_1,\ldots,v_{M-1},s_{M-1}\right).
\end{equation}
Note that if a tree model contains $M$ terminal nodes, it consequently has $M-1$ internal splits. These internal nodes $r\in\{1,\ldots,M-1\}$ consist of a splitting variable $v_r\in\{1,\ldots,p\}$ and the associated split rule (or point) $s_r$. For ordered and numeric variables $X_{v_r}$, the split rule $s_r$ involves a cutoff and there are $u-1$ possible splits if $X_{v_r}$ takes $u$ distinct values. For a categorical variable with $k$ levels, the split rule contains a (non-empty) subset of $\{1,\ldots,k\}$ and there are $2^{k-1}-1$ possible splits. Based on the split rules, observations are sent to either the first or second subset. The product of these combinations forms all potential elements $\theta$ from $\Theta_M$, the space of conceivable trees with $M$ terminal nodes. The overall parameter space is then $\Theta=\bigcup_{M=1}^{M_{max}}\Theta_M$.

Let $\{(\xb_i,y_i)\}^N_{i=1}$ denote the $N$ observed predictor-response pairs in the data set, where $y_i\in\{0,1\}$ describes the response and $\xb_i=\left(x_{i1},\ldots,x_{ip}\right)^t$ represents the $p$ associated predictor variables of instance $i$. For each customer, a probability estimate or score, $s\in [0,1]$, can be calculated based on the observed features $\xb_i$ of the customer.
In the churn context, the probability of an instance being a churner is determined by a score function $s\left(\mathbf{X,\theta}\right)$ which is based on all explanatory variables $\mathbf{X}$ and the chosen tree structure $\theta$ from (\ref{tree}). The instances from class $0$ (no churn) are assumed to have a lower score than the instances from class $1$ (churn). We define the score of instance $i$ as
\begin{equation}\label{score}
s(\mathbf{x}_i,\theta)=\sum_{m=1}^{|\theta|}\hat{p}_{m}I(\xb_i\in R_m)
\end{equation}
where $|\theta|$ is the number of terminal nodes (i.e. regions $R_m$) and
\begin{equation}\label{pm}
\hat{p}_{m}=\frac{1}{N_m}\sum_{i:\xb_i\in R_m}I(y_i=1)
\end{equation}
is the proportion of class $1$ observations in node $m$ which represents a region $R_m$ with $N_m$ observations. It is obvious from (\ref{score}) and (\ref{pm}) that the churn scores lie between zero and one. Note that a higher score indicates a higher likelihood of churning. These scores are often converted to predicted classes $\hat{y}\in\{0,1\}$ by comparing them with a classification threshold $t\in[0,1]$. All instances with a score $s$ smaller than $t$ are classified as non-churners (class $0$), whereas instances with $s$ larger than $t$ are classified as churners (class $1$):
\begin{align}\label{classification}
\begin{split}
s(\mathbf{x}_i,\theta)\leq t &\Rightarrow \hat{y}_i=0 \\
s(\mathbf{x}_i,\theta)> t &\Rightarrow \hat{y}_i=1
\end{split}
\end{align}

\subsection{General Fitness Function of Decision Trees\label{subsec:generalfitnessfunction}}

At the heart of most classification methods lies a fitness function that is optimized by the respective algorithm. In order to prevent that the estimated model is overfitted on the training sample, the complexity of the tree is often included in the algorithm's fitness function. As done by \citet{breiman1984classification} and \citet{grubinger2011evtree}, we measure the complexity of a tree by a function of the number of terminal nodes, $|\theta|$, without further considering the depth or the shape of the tree. The aim of the algorithm is then to find the classification tree which optimizes a given fitness function which describes some tradeoff between prediction performance and complexity \citep{grubinger2011evtree}:
\begin{equation}\label{thetahat}
\hat{\theta}=\underset{\theta\in\Theta}{\operatorname{argmin}}\text{ loss}\{Y,s\left(\mathbf{X},\theta\right)\}+\text{comp}(\theta)
\end{equation}
where $\text{loss}(\cdot,\cdot)$ represents a suitable loss function for $Y$. Popular loss functions for classification are the misclassification rate, deviance (i.e. cross-entropy) or Gini index. The function $\text{comp}(\cdot)$ is monotonically non-decreasing in the number of terminal nodes $|\theta|=M$ of the tree $\theta$. As a result, more complex models are penalized in the tree selection process as they are less favorable. Note that finding $\hat{\theta}$ requires a search over all spaces $\Theta_M$ for $M\in\{1,\ldots,M_{max}\}$.

Even for medium sized problems with a relatively small number of observations and features, it is clear that the complete parameter space $\Theta$ can become very large. Instead of searching all possible combinations in $\Theta$ simultaneously and measuring their fitness, classic recursive partitioning algorithms, like CART \citep{breiman1984classification} and C4.5 \citep{quinlan1993c45}, only consider one split at a time. This means that at each internal node $r\in\{1,\ldots,M-1\}$, the split variable $v_r$ and the corresponding split point $s_r$ are selected to locally minimize the loss function. Forward-search recursive partitioning methods only search each tuple $(v_r,s_r)$ once without taking the subsequent split rules into account. Although it has been shown that this approach is an efficient heuristic, it typically leads to a local optimal solution. An alternative way to search over the parameter space of trees is to use global optimization methods like an evolutionary algorithm as is done by \citet{grubinger2011evtree}.

\subsection{Profit-based Classification Performance Evaluation\label{subsec:performancemeasures}}

The quality of a churn model is traditionally assessed as a binary classification model, using a performance measure. Most performance metrics are extracted from a confusion matrix as shown in Table \ref{confusionmatrix}.  Depending on a given classification threshold $t\in[0,1]$, the confusion matrix tabulates the numbers of correct and incorrect classifications based on the churn scores $s$ produced by the predictive model. If $s\leq t$, the model  assigns a ``no churn'' label to the instance, and if $s>t$, the model assigns a ``churn'' label.
\begin{table}[h]
	\begin{center}
		\begin{tabular}{c c c}
			\hline
			\textbf{Predicted class} & \multicolumn{2}{c}{\textbf{Actual class}} \\ \cline{2-3}
			& Class $0$ & Class $1$ \\ \hline
			Class $0$ & $\pi_0F_0(t)N$ & $\pi_1F_1(t)N$ \\
			& $[b_0=c(0\text{ }|\text{ }0)]$ & $[c_1=c(0\text{ }|\text{ }1)]$ \\
			& & \\
			Class $1$ & $\pi_0(1-F_0(t))N$ & $\pi_1(1-F_1(t))N$ \\
			& $[c_0=c(1\text{ }|\text{ }0)]$ & $[b_1=c(1\text{ }|\text{ }1)]$ \\
			\hline
		\end{tabular}
	\end{center}
	\caption{Confusion matrix with associated benefits ($b_k$) and costs ($c_k$), $k\in\{0,1\}$, for a correct and incorrect classification, respectively.}
	\label{confusionmatrix}
\end{table}

The prior class probabilities of instances belonging to class $0$ or $1$ are represented by $\pi_0$ and $\pi_1$, respectively. Furthermore, $f_0(s)$ and $f_1(s)$ are the probability density functions of the classification scores, whereas $F_0(s)$ and $F_1(s)$ are the cumulative distribution functions (CDFs) of the scores for class $0$ and $1$, respectively. The confusion matrix will of course look differently depending on the chosen classification threshold $t\in[0,1]$. Examples of some well accepted measures for binary classification problems are:
\begin{align}
\textit{Accuracy}(t) &= \pi_0F_0(t)+\pi_1\left(1-F_1(t)\right) \label{eq:accuracy} \\
\textit{Error rate}(t) &= 1 - \textit{Accuracy}(t) = \pi_0\left(1-F_0(t)\right) + \pi_1F_1(t) \label{eq:errorrate} \\
\textit{Recall}(t) &= F_0(t) \label{eq:recall} \\
\textit{Precision}(t) &= \pi_0F_0(t) / \left(\pi_0F_0(t)+\pi_1F_1(t)\right) \label{eq:precision} \\
F_1\textit{ measure}(t) &= 2\pi_0F_0(t)/\left(\pi_0+\pi_0F_0(t)+\pi_1F_1(t)\right) \label{eq:F1measure} \\
\textit{MER} &= \min_{\forall t}\{Error\text{ }rate(t)\} \label{eq:MER} \\
\textit{AUC} &= \int_{-\infty}^{+\infty}F_0(s)f_1(s)ds \label{eq:AUC}
\end{align}
Most of these performance measures are a function of the classification threshold. However, the area under the ROC curve (AUC) and the minimum error rate (MER) do not require to specify the classification threshold which makes them one of the most employed measures to objectively evaluate the classification performance. The AUC of a classifier can be interpreted as being the probability that a randomly chosen churner is predicted a higher score than a randomly chosen non-churner. Therefore, a higher AUC indicates superior classification performance.

The outcome of a classification task is typically used as input to the retention campaign which leads to costs for misclassifications and benefits for correct classifications. The cost or benefit related to labeling an instance from class $k$ as a class $l$ instance is denoted with $c(l|k)$, $k,l\in\{0,1\}$. As indicated in Table \ref{confusionmatrix}, correct classifications are rewarded with a benefit $b_k=c(l=k|k)$ while misclassifications are penalized and associated with costs $c_k=c(l\neq k|k)$. Both costs and benefits will by convention be non-negative, unless stated otherwise, and thus a minus sign will be applied for misclassification costs. In general, the average classification profit of a classifier is then computed by offsetting the total cost and benefit against each other, and dividing by the number of instances \citep{verbraken2013novel}:
\begin{equation}\label{eq:generalaverageprofit}
P(t;b_0,c_0,b_1,c_1)=b_0\pi_0F_0(t)+b_1\pi_1(1-F_1(t))-c_0\pi_0(1-F_0(t))-c_1\pi_1F_1(t)
\end{equation}
Note that the average profit is a function of the classification threshold $t$. From (\ref{eq:accuracy})$-$(\ref{eq:AUC}), it can be seen that none of the listed performance measures explicitly take classification costs or benefits into account. Therefore, they are only valid if the gains of correct classifications and the severities of misclassifications are equal. Churn prediction problems are typically dealing with high class imbalance where the minority class (i.e. churners) is of primary interest, thus making the assumption of equal benefits and costs unrealistic. Indeed, misclassifying an actual churner as a non-churner, also known as a false negative, results in the loss of a customer, while misclassifying a non-churner as a churner, also known as a false postive, leads to additional costs to the company (e.g. cost of contacting the customer and the cost of an incentive offer) since those misclassified customers do not intend to leave. As a result, applying accuracy-based performance measures for the evaluation and the selection of a classifier in churn management is not recommended. 

Accordingly, \citet{verbraken2013novel} proposed the cost benefit analysis framework for customer churn, which incorporates the costs associated with a retention campaign and the benefits of retained customers. This logic reformulates the general definition (\ref{eq:generalaverageprofit}) specifically for churn management campaigns.  As a result, the average classification profit of a classifier for customer churn is defined by \citep{verbraken2013novel}:
\begin{equation}\label{eq:churnaverageprofit}
P_C(t;\gamma,CLV,\delta,\phi)=CLV\left(\gamma(1-\delta)-\phi\right)\pi_0F_0(t)-CLV(\delta+\phi)\pi_1F_1(t),
\end{equation}
where $t$ is the classification threshold and $\gamma$ is the probability that a targeted would-be churner accepts a special offer and remains a customer. $CLV$ represents the constant customer lifetime value per retained customer (e.g. \euro $200$). The two dimensionless parameters $\delta=d/CLV$ and $\phi=f/CLV$ are derived from $d$, the constant cost of the retention offer (e.g. \euro $10$), and $f$, the constant cost of contacting a customer (e.g. \euro $1$). It is assumed that all costs involved are strictly positive and $CLV>d$. The values between brackets are the recommended default values for customer retention campaigns in the telecommunication sector as suggested by \citet{verbraken2013novel}. Of course, the classification threshold is chosen such that it maximizes the average profit:
\begin{equation}\label{optimalthreshold}
t_{opt}=\underset{\forall t}{\operatorname{argmax}}\left\{P_C(t;\gamma,CLV,\delta,\phi)\right\}
\end{equation}
assuming that all values of $\gamma$, $CLV$, $\delta$ and $\phi$ are known. The proposed cost benefit framework comprises a deterministic and a probabilistic profit-based performance measure. The former is the \textit{maximum profit measure for customer churn} (MPC) \citep{verbraken2013novel}:
\begin{equation}\label{MPC}
MPC=\max_{\forall t}\left\{P_C(t;\gamma,CLV,\delta,\phi)\right\}
\end{equation}
The MPC is deterministic in the sense that all parameters related to costs and benefits ($\gamma$, $CLV$, $\delta$ and $\phi$) are assumed to be known. In a retention campaign, most of these parameters can be approximated with sufficient precision, except the probability $\gamma$ that a customer will accept the offer. This can be solved by assigning a probability density function to $\gamma$, denoted as $h(\gamma)$, which yields the \textit{expected maximum profit measure for customer churn} (EMPC):
\begin{equation}\label{EMPC}
EMPC=\int_\gamma P_C(t_{opt}(\gamma);\gamma,CLV,\delta,\phi)h(\gamma)\textrm{d}\gamma,
\end{equation}
\citet{verbraken2013novel} decided to specify a beta distribution for $h(\gamma)$ with the restrictions that its parameters are $\alpha'>1$ and $\beta'>1$. Furthermore, the authors propose to set $\alpha'$ and $\beta'$ equal to $6$ and $14$, respectively. The EMPC follows a probabilistic approach that considers a range of $\gamma$ values, representing the uncertainty in that parameter. In this paper, we will focus on the EMPC because this scenario is more likely to be encountered in practice. Yet, the most profitable classifier can be identitief unambiguously by both profit measures.

Additionally, the proposed cost benefit framework provides the \textit{(expected) profit maximizing fraction for customer churn}, $\overline{\eta}$, which gives practitioners an estimate of the optimal fraction of the customer base which needs to be targeted in the retention campaign.  Following the deterministic approach, it becomes \citep{verbraken2013novel}:
\begin{equation}\label{mpcfraction}
\overline{\eta}_{mpc}=\pi_0F_0(t_{opt})+\pi_1F_1(t_{opt})
\end{equation}
whereas the profit maximizing fraction derived from the EMPC framework is defined as \citep{verbraken2013novel}:
\begin{equation}\label{empcfraction}
\overline{\eta}_{empc}=\int_{\gamma}\left[\pi_0F_0(t_{opt}(\gamma))+\pi_1F_1(t_{opt}(\gamma))\right]h(\gamma)\textrm{d}\gamma
\end{equation}
Making an arbitrary choice of taking, for example, the top $10\%$ of predicted would-be churners, will likely result in suboptimal profits. Therefore, the $\overline{\eta}$ estimates are appealing to practitioners since they help to determine how many customers should be targeted in the retention campaign for attaining maximal profit.

\subsection{Evolutionary algorithms\label{subsec:evolutionaryalgorithms}}

An evolutionary algorithm (EA) is a metaheuristic algorithm inspired by the biological process of evolution that allows solving complex optimization problems \citep{eiben2015introduction, yu2010introduction}.
Over the years, many different variants of EA have been proposed within the domain of evolutionary computation with the most prominent techniques being genetic algorithms \citep{holland1992adaptation}, evolution strategies \citep{rechenberg1973evostrat}, evolutionary programming \citep{fogel1967artifical}, and genetic programming \citep{koza1992genetic}.
Despite the technical differences, the same fundamental idea is behind all evolutionary methods \citep{eiben2015introduction}.

Within some defined environment, a population of individuals undergoes an evolution process that is guided by some fitness function, a measure for judging the quality of an individual, with the goal to find the fittest individual within that environment.
In this evolution process, the population members compete against each other for survival, where each individual represents a candidate solution to the problem at hand.
In the spirit of \emph{survival of the fittest}, the evolutionary system is typically designed such that individuals with a higher fitness have a higher chance to survive.
The fittest individual returned by the EA represents ultimately the final solution to the optimization problem.

At the start, one usually creates the population, the set of candidate solutions, in a random fashion and then repeatedly applies (a subset of) genetic operators to evolve the population toward the optimal solution.
The most distinguished genetic operators are \emph{selection}, \emph{crossover}, and \emph{mutation}.
As the name suggests, the selection operator chooses (either deterministically or stochastically) individuals based on their fitness values that should undergo a genetic operation.
Crossover is the recombination operator that is applied to two or more selected candidates (the so-called parents), which as a result produces one or more new candidates (the children) \citep{eiben2015introduction}.
The inner values (called genes) of the parents are thereby exchanged in a predefined manner.
The mutation operator is another way of creating a new candidate from one selected candidate, usually by randomly changing the genes.
By repeatedly applying the genetic operators and updating the population, the EA converges toward the optimal solution from generation to generation, assuming it does not stop prematurely.

The immense flexibility of designing the building blocks of an EA system enables users to solve highly complex optimization problems.
Assuming the building blocks are properly defined and no premature termination has occurred, EAs are capable of handling complex data structures such as classification and regression trees and return globally optimal solutions \citep{grubinger2011evtree}.
It should therefore come as no surprise that thanks to this property EAs are often the preferred choice and therefore have been applied to a large variety of optimization and search problems (with respect to data mining, see, for example, \citet{freitas2003survey}).

\subsection{Related work\label{subsec:relatedwork}}

The idea of directly integrating the EMPC as a profit measure into the model construction is also considered by \citet{stripling2015profit, stripling2017journal}. Here, a classifier, called ProfLogit, is proposed that maximizes the EMPC directly in the training step using a genetic algorithm. The internal model structure of their classifier resembles a lasso-regularized logistic model. Besides the cost benefit framework of the EMPC, there are other frameworks that are used for customer churn predictive modeling. The framework of the EMPC, as proposed by \citet{verbraken2013novel}, allows to allocate costs and benefits to each particular class. An alternative approach is to measure classification costs at the level of individual customers. This means that distinct costs are set for each respective customer which leads to a more detailed description of how costly the missclassification of a particular customer is. The classification performance of a classifier can then be measured by the total cost as the sum of all individual costs. This example-dependent cost-sensitive framework for churn is proposed by \citet{bahnsen2015novel}. Their framework is purely cost-based since it does not account for any benefits coming from retained customers. For instance, unlike in the cost benefit framework of the EMPC, the customer lifetime value (CLV) is considered as a cost for effectively churned customers rather than a benefit of retained customers. In \citet{bahnsen2015example}, the example-dependent cost-sensitive framework is introduced to decision trees. Their algorithm is based on a recursive partitioning approach which builds a decision tree by optimizing a cost based impurity measure that incorporates the different example-dependent costs as well as cost based pruning criteria. In \citet{glady2009modeling}, the concept of a churner is defined as a function of the customer lifetime value. More specifically, a churner is defined as a customer whose CLV is decreasing where CLV is defined as the discounted value of future marginal earnings related to the customer's activity. Additionally, a new loss function is introduced to assess the misclassification cost of a customer that is incurred by the decrease in CLV.

The extensive literature review given by \citet{verbeke2011building} shows that predictive classification techniques for customer churn are increasingly researched. Various machine learning techniques have been set up for the task of predicting customer churn, including support vector machines \citep{chen2012hierarchical} and ensemble methodes \citep{van2007improved}. In saturated markets, the attraction of new customers is very challenging and costs approximately five to six times more than preventing existing customers from churning \citep{athanassopoulos2000customer}. One example of such a market is the telephone service industry. As a consequence, multiple data mining techniqes have been researched and applied within the telco industry \citep{verbeke2012new}.

\section{Profit Maximizing Classification Trees\label{sec:ProfTree}}

In this section, we introduce our new ProfTree classification technique, which optimizes the splitting variables and associated split rules according to the EMPC, aiming to produce the most profitable classifier. An evolutionary algorithm is used to find the optimal tree model that corresponds to a maximum on the EMPC landscape. In what follows we describe  the building blocks of ProfTree.

\subsection{Fitness Function of ProfTree\label{subsec:ObjectiveFunctionProfTree}}

The fitness function represents the requirements to which the population of trees should adapt. In general, these requirements are formulated by (\ref{thetahat}). ProfTree's fitness function is defined by substituting the loss function in (\ref{thetahat}) with the EMPC measure (\ref{EMPC}) and measuring the complexity of a tree as its number of terminal nodes, weighted by a user-specified parameter $\lambda$. This ultimately yields a profit-sensitive classification model:
\begin{equation}\label{thetahatempc}
\hat{\theta}^{empc}=\underset{\theta\in\Theta}{\operatorname{argmax}}\text{ }EMPC(\theta)-\lambda\cdot|\theta|
\end{equation}
where $\hat{\theta}^{empc}$ is the globally optimal classification tree according to the EMPC measure with minimized complexity. The underlying model structure of ProfTree is based on a partitioning algorithm. However, rather then using a classic recursive algorithm (like CART or C4.5), the tree $\theta$ is optimized for maximum profit through an evolutionary algorithm. In \citet{grubinger2011evtree}, the \textit{evtree} package is described, which implements an evolutionary algorithm for learning globally optimal classification trees in \verb|R|. ProfTree is also implemented in \verb|R| and it exploits the functionalitities available in the \textit{evtree} package \citep{grubinger2011evtree} and the \textit{EMP} package \citep{bravo2015emp}.

\subsection{The ProfTree algorithm\label{subsec:ProfTreeAlgorithm}}

The evolutionary algorithm behind ProfTree is largely based on the implementation available in the \textit{evtree} package \citep{grubinger2011evtree}. The algorithm starts by initializing a population of randomly created trees. Each initial tree of the population is generated by adding a split rule to the root node in which both the splitting variable and corresponding split point are selected randomly. These trees are considered to be the first generation.

Once the population is generated, each individual tree is evaluated by the fitness function $EMPC(\theta)-\lambda\cdot|\theta|$ for a given value of the parameter $\lambda$. In order to improve the fitness of the trees, each of them are altered via so-called variation operators. These operators cause the trees to grow according to (\ref{thetahatempc}) in order to maximize their fitness. After evaluating each new solution and comparing it with the previous version, the algorithm will select the fittest trees as the survivors for the next generation. Once a new generation has been created from these survivors, the iteration starts again by evaluating the quality of each new tree in the population. The overall fitness of the population will increase by each iteration until a termination condition is met. The details of the termination conditions are given below. The result of the ProfTree algorithm is the tree with the highest fitness according to (\ref{thetahatempc}).

\subsection{Variation operators\label{subsec:VariationOperators}}

In each iteration, each tree in the population is modified by one of the variation operators which consist of four types of mutation operators and one crossover operator. The details of these operatos are described by \citet{grubinger2011evtree}.

\begin{description}
	\item[Split:] The split operator adds a randomly generated split rule to a randomly selected terminal node.
	\item[Prune:] The prune operator prunes a random internal node, which has two terminal nodes as successors, into a terminal node.
	\item[Major split rule mutation:] This operator selects a random internal node and changes the entire split rule by altering both the corresponding split variable and the split point.
	\item[Minor split rule mutation:] This operator is similar to the previous one, except that it does not change the split variable; only the split point is slightly altered.
	\item[Crossover:] The crossover operator randomly choses two ``parent'' trees and exchanges randomly selected subtrees between them. Due to this operator, some trees are selected more than once in each iteration as it needs a second parent.
\end{description}

\subsection{Termination criteria\label{subsec:TerminationCriteria}}

The same termination criteria of the evolutionary search are used as the ones from \textit{evtree} \citep{grubinger2011evtree}. The ProfTree algorithm executes a minimum of $1000$ iterations after which it terminates when the quality of the best $5\%$ of trees has not improved for $100$ iterations. If this convergence criterium is not met, the algorithm will terminate after a user-specified number of iterations, which is set at $10\,000$ by default. The algorithm returns the tree with the highest performance according to (\ref{thetahatempc}).

\subsection{Control parameters\label{subsec:ControlParameters}}

The implementation of ProfTree contains several parameters that control the evolutionary search. The solution can be constrained to a minimum number of observations in each internal node, a minimum number of observations in each terminal node, and a maximum tree depth. The parameter $\lambda$ controls the overall complexity (i.e. number of terminal nodes) of the trees in each generation. If $\lambda$ is increased, the number of terminal nodes will decrease and vice versa. The evolutionary search can further be controlled by specifying the number of trees in the population (which is set at $100$ by default) as well as the probabilities for the variation operators. In each modification step, one of these variation operators is chosen at random for each tree. The probabilities by which the operators are selected can be specified by the user. By default, each of the five variation operators are given a $20\%$ probability of being selected.

\subsection{Tuning the Regularization Parameter\label{subsec:tuninglambda}}

The parameter $\lambda$ governs the tradeoff between (expected) profit and tree size. In order to estimate the optimal regularization parameter value, $\lambda_{\text{opt}}$, we generate a grid of $\lambda$ values as follows:
\begin{equation}\label{lambda_grid}
\Lambda = \{\lambda\text{ }|\text{ }\lambda_{\text{min}} < \lambda < \lambda_{\text{max}}\}.
\end{equation}
The number of grid values is an arbitrary choice, but it should be large enough to create a dense grid. Based on many preliminary studies on data sets with widely different dimensions, we found that the optimal value $\lambda_{\text{opt}}$ typically resides in the interval $[0,1]$. The estimated optimal value $\lambda_{opt}$ corresponds to the value on the grid with the highest EMPC performance on a validation set. In order to confidently determine $\lambda_{opt}$, a reliable performance estimate has to be obtained for each $\lambda$ value of the grid. The underlying evolutionary algorithm of ProfTree, however, is intrinsicly stochastic in nature. This requires that the analysis has to be repeated several times to obtain average performance estimates. Therefore, we perform five replications of twofold cross-validation ($5\times2$ cv) \citep{dietterich1998approximate, demvsar2006statistical} in which each fold is stratified according to the churn indicator. In each replication, ProfTree is trained with a given $\lambda$ on each fold, and its EMPC performance is measured on the other fold. Next, the average of the $10$ estimates becomes the associated performance estimate for the given $\lambda$. This procedure is performed for all $\lambda\in\Lambda$, and $\lambda_{opt}$ corresponds to the value with the highest average EMPC performance.

\subsection{Performance Measures based on the Expected Profit Maximizing Fraction for Customer Churn\label{subsec:etaperformancemeasures}}

Besides the profit measures MPC and EMPC, we present three additional performance measures that are based on the expected profit maximizing fraction (\ref{empcfraction}). These measures were first introduced by \citet{stripling2017journal} and are related to the notion of precision, recall, and the $F_1$ measure, but they are independent of the classification threshold $t$. We will state the definitions of these three measures below.

The expected profit maximizing fraction for customer churn, $\overline{\eta}_{empc}$, tells us how many customers to target in the retention campaign in order to achieve maximum profit. To do so, all customers are ordered according to their predicted churn score. Next, $\overline{\eta}_{empc}$ indicates which top fraction of the predicted would-be churners should be targeted in the campaign. This is especially interesting for practioners, since subjectively choosing the cutoff fraction is avoided. The result is a customer list with the top would-be churners. The three additional performance measures are based on this list of customers \citep{stripling2017journal}. The $\overline{\eta}$-precision (or hit rate) for customer churn, $\overline{\eta}_p$, is the proportion of correct identifications of would-be churners in the $\overline{\eta}_{empc}$-based customer list. The $\overline{\eta}$-recall for customer churn, $\overline{\eta}_r$, is the proportion of churners that is included in the $\overline{\eta}_{empc}$-based customer list. Precision and recall are often combined into the $F_1$ measure, which represents the harmonic mean between the two measures. The $F_1$ measure reaches its best value at 1 and worst at 0. The $\overline{\eta}$-based $F_1$ measure for customer churn, $\overline{\eta}_{F_1}$, is defined as:
\begin{equation}\label{etaF1}
\overline{\eta}_{F_1} = 2\frac{\overline{\eta}_p\overline{\eta}_r}{\overline{\eta}_p+\overline{\eta}_r}
\end{equation}
For all three measures, higher values indicate higher effectiveness of the customer list and thus better performance of the classifier. However, since these measures calculate accuracy only, the final model should be selected based on the EMPC as we aim for maximum profit. Nevertheless, it is interesting to compare hit rate ($\overline{\eta}_p$) among classifiers in order to assess their effectiveness of correctly identifying churners.

\section{Case study \label{sec:casestudy}}

In this section, we will illustrate the benefit of using the ProfTree classifier on a customer churn data set from a South Korean telecom operator. The data set contains a sample of 889 customers and 10 explanatory variables (see Table \ref{KoreanPersonalVariables}); 277 of these customers (i.e. $31.16\%$) were recorded as churners.
\begin{table}[h]
	\begin{center}
		\begin{tabular}{l l}
			\hline
			Variable & Description \\ \hline
			\verb|churn| & Did the customer leave the telecom operator? \\
			\verb|region| & Region where the customer lives.\\
			\verb|prod_num| & Number that identifies the customer's product.\\
			\verb|active_months| & Time since the customer joined the operator (in months).\\
			\verb|contract_period| & Length of the contract period.\\
			\verb|revenue_avg| & Average revenue.\\
			\verb|nonpay_period| & How long did the customer not pay the bills?\\
			\verb|overdue_amt| & Amount that the customer is overdue.\\
			\verb|count_disconnect| & Number of times the service was disconnected.\\
			\verb|count_complaint| & Number of filed complaints.\\
			\verb|autopay| & Did the customer use the automatic payment option?\\ \hline
		\end{tabular}
	\end{center}
	\caption{Variables of the Korean telecom churn data.}
	\label{KoreanPersonalVariables}
\end{table}

The goal is to build a model that predicts would-be churners while taking profitability into account. Moreover, we want the model to be easily interpretable such that the results can be communicated to the marketing departement. Therefore, we will build a tree model with a maximal depth of three levels of splits. Besides ProfTree, we use other classification tree methods like EvTree, CART and conditional inference trees. EvTree \citep{grubinger2011evtree} uses an evolutionary algorithm for learning globally optimal classification trees while CART \citep{breiman1984classification} is a commonly used classification tree algorithm which uses a greedy heuristic approach, where split rules are selected in a forward stepwise search for recursively partitioning the data into groups. The \textit{ctree} algorithm \citep{hothorn2006unbiased} builds a conditional inference tree which estimates a regression relationship by binary recursive partitioning in a conditional inference framework. All trees are constrained to have a minimum of 10 observations per terminal node, 20 observations per internal node, and a maximum tree depth of 3. Furthermore, the conditional inference tree is constructed with a significance level of $1\%$ rather than the default $5\%$ level since it seems more appropriate for $889$ observations. Another well-known recursive partitioning method is C4.5 \citep{quinlan1993c45}. However, the results for C4.5 on this case study are not reported because the tree depth can not be restricted by the method's implementation.

\begin{figure}
	\centering
	\begin{subfigure}[b]{0.95\textwidth}
		\includegraphics[width=1\linewidth]{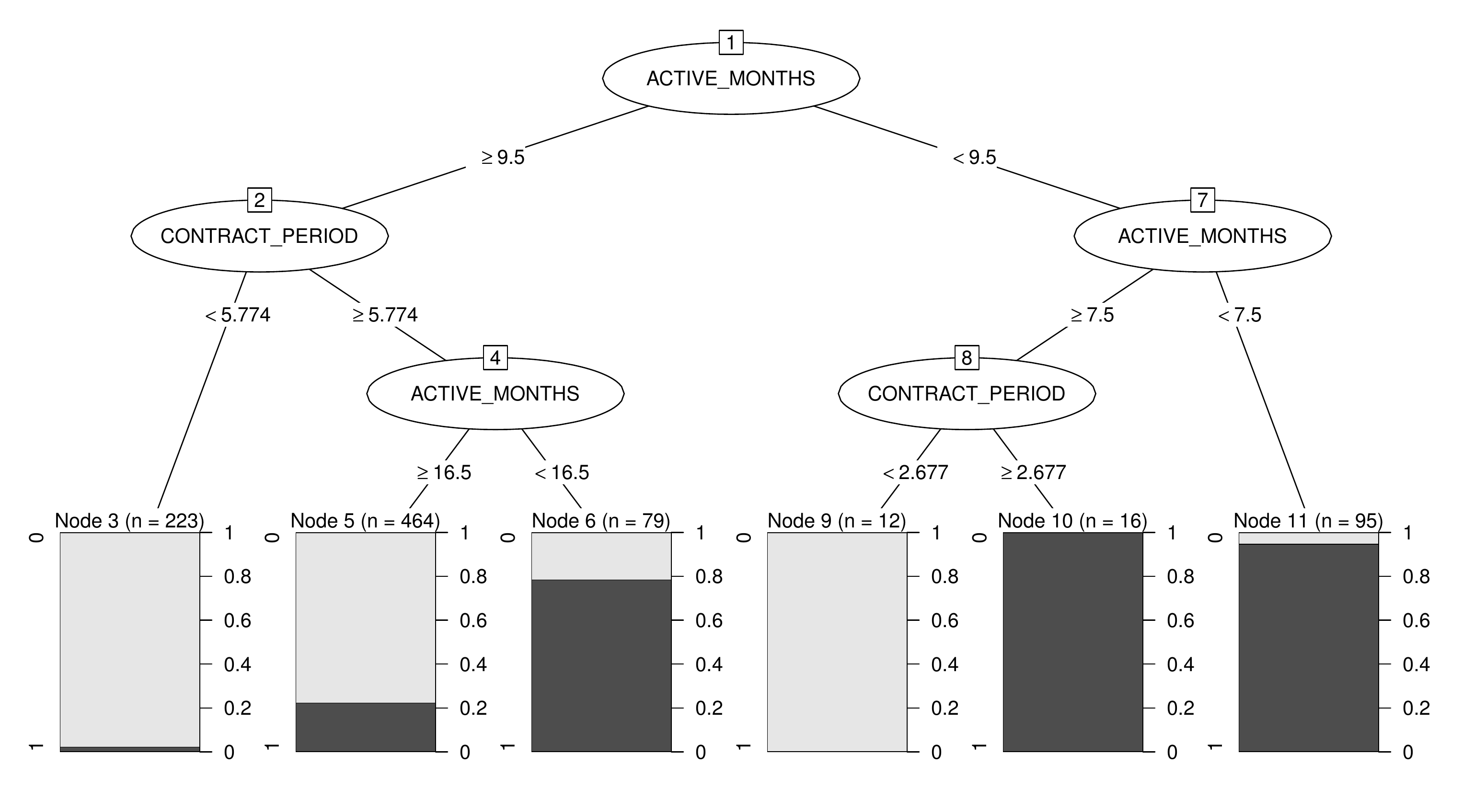}
		\caption{}
		\label{fig:CART_Koreanpersonal_casestudy}
	\end{subfigure}
	\begin{subfigure}[b]{0.95\textwidth}
		\includegraphics[width=1\linewidth]{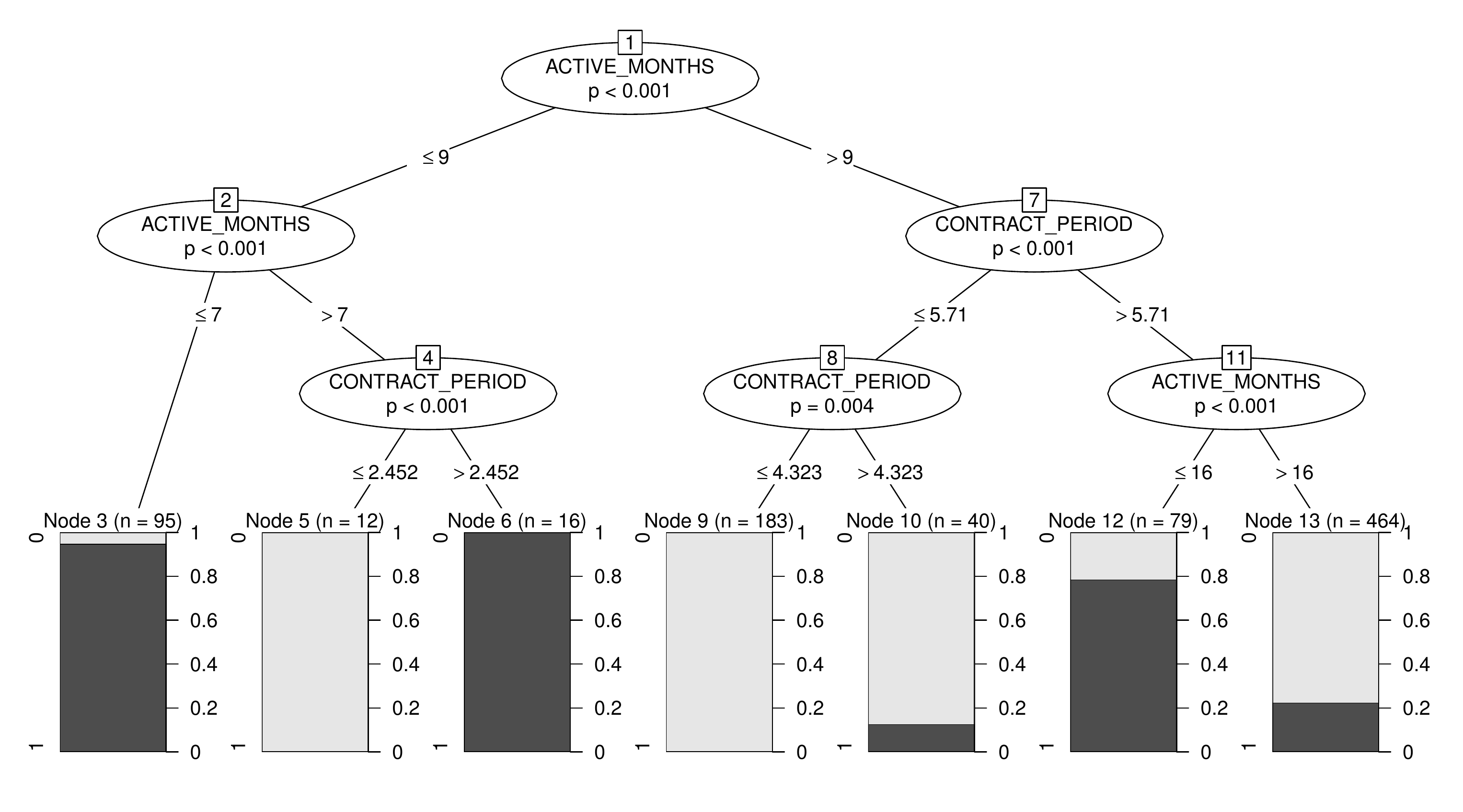}
		\caption{}
		\label{fig:Ctree_Koreanpersonal_casestudy}
	\end{subfigure}
	\caption[Growing CART and ctree]{Trees for customer churn prediction constructed by CART (a) and ctree (b).}
	\label{fig:CART_Ctree_Koreanpersonal_casestudy}
\end{figure}

First, we grow the forward-search trees with CART and ctree in Figure \ref{fig:CART_Ctree_Koreanpersonal_casestudy}.
Despite the fact that the data set contains 10 explanatory variables, the CART and ctree algorithm only use the variables \verb|active_months| and \verb|contract_period| to predict the churning propensity of the customers.

\begin{figure}
	\centering
	\begin{subfigure}[b]{0.75\textwidth}
		\includegraphics[width=1\linewidth]{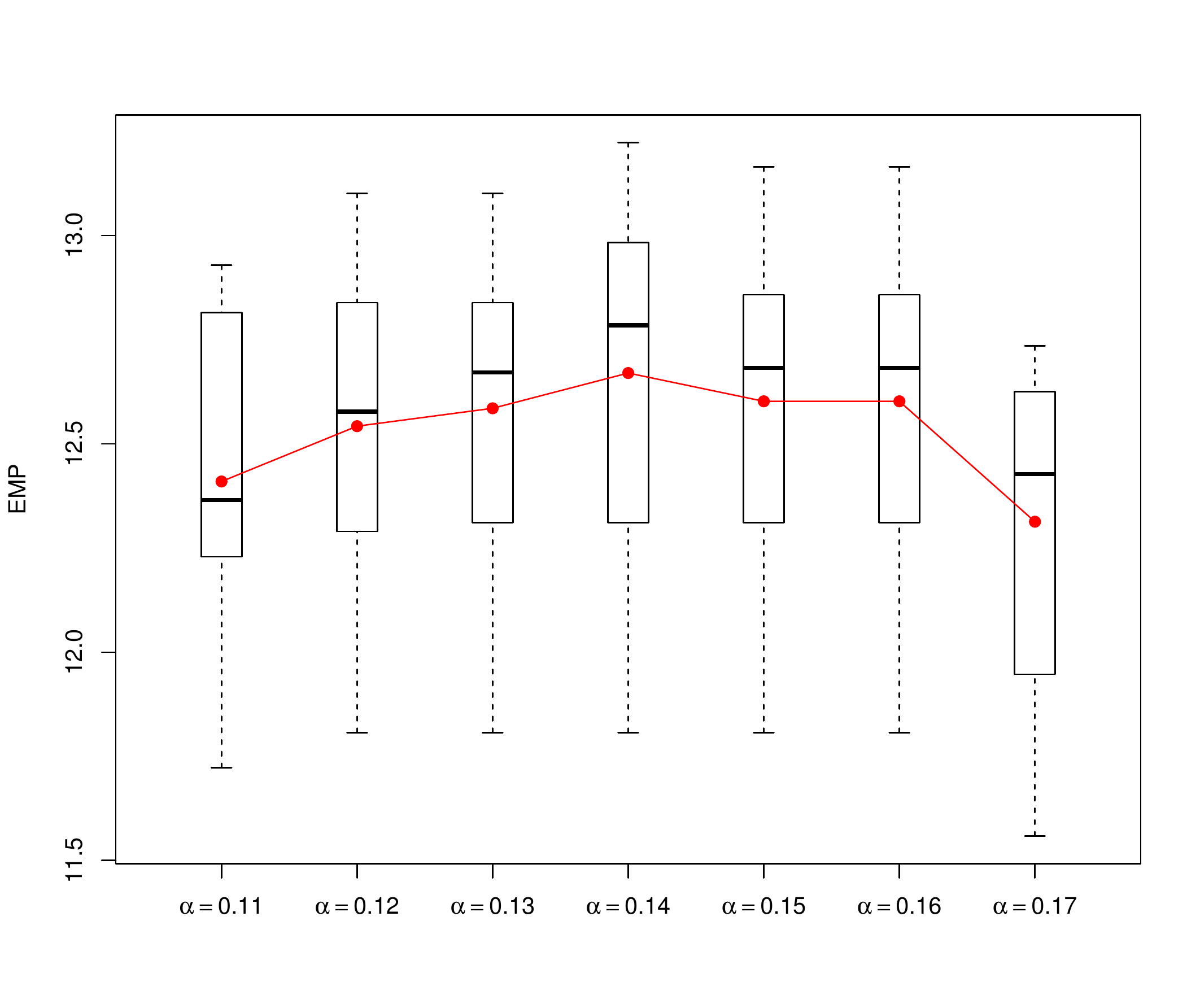}
		\caption{}
		\label{fig:EvTree_tuningalpha_casestudy} 
	\end{subfigure}
	\begin{subfigure}[b]{0.75\textwidth}
		\includegraphics[width=1\linewidth]{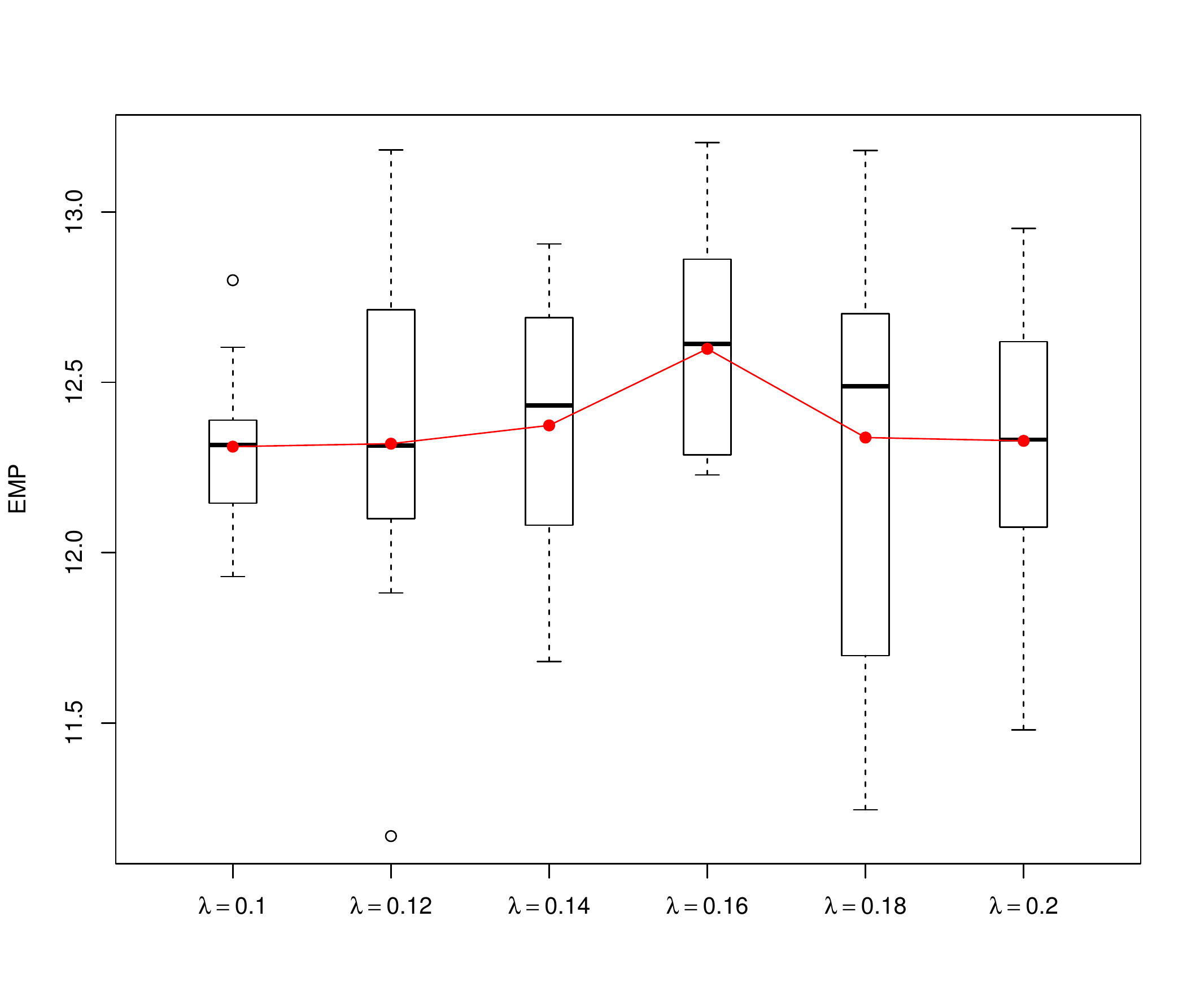}
		\caption{}
		\label{fig:ProfTree_tuninglambda_casestudy}
	\end{subfigure}
	\caption[Tuning regularization parameters]{Average EMPC performance (red dot) for various values of EvTree's $\alpha$ (a) and ProfTree's $\lambda$ (b).}
\end{figure}

\begin{figure}
	\centering
	\begin{subfigure}[b]{0.95\textwidth}
		\includegraphics[width=1\linewidth]{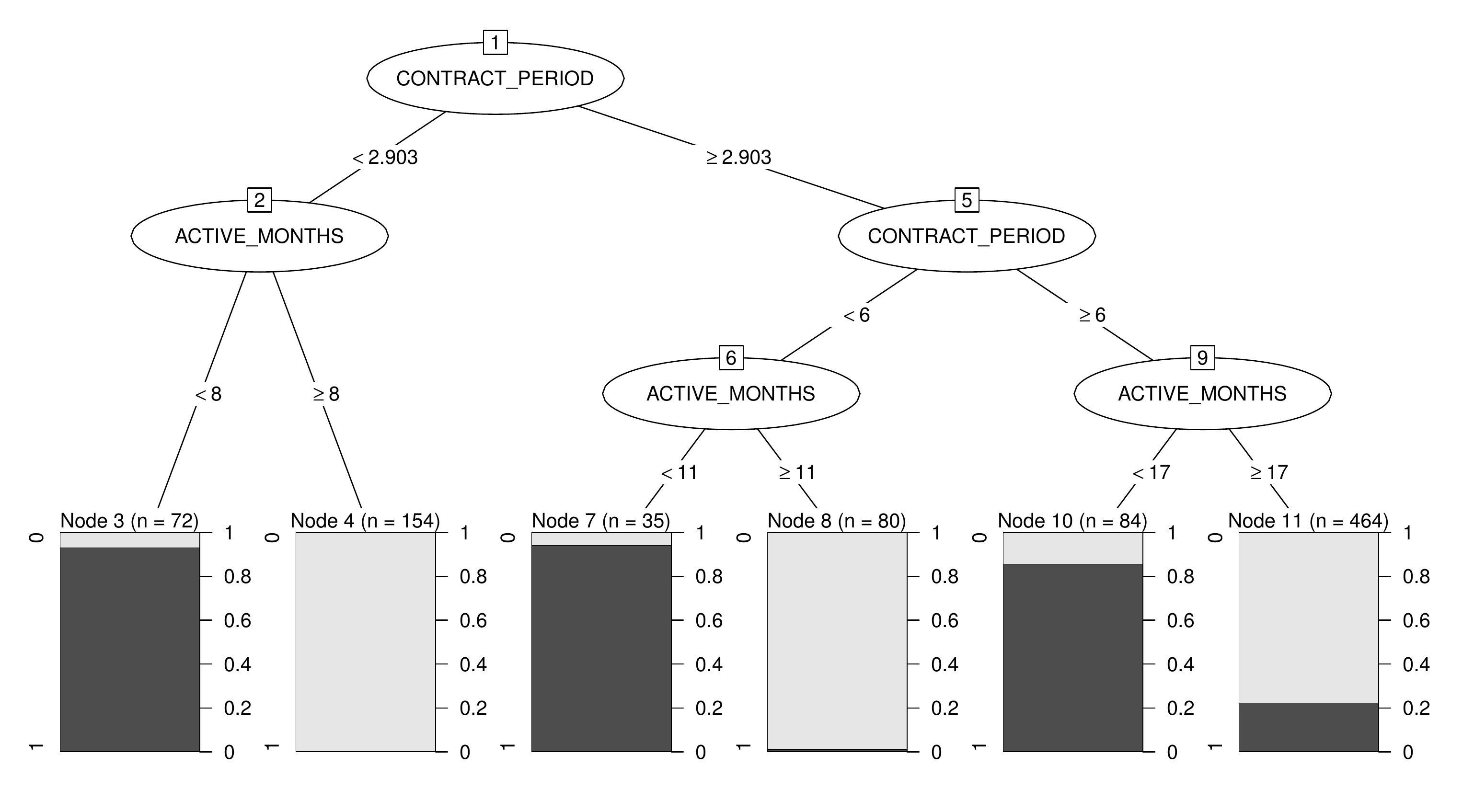}
		\caption{}
		\label{fig:EvTree_Koreanpersonal_casestudy} 
	\end{subfigure}
	\begin{subfigure}[b]{0.95\textwidth}
		\includegraphics[width=1\linewidth]{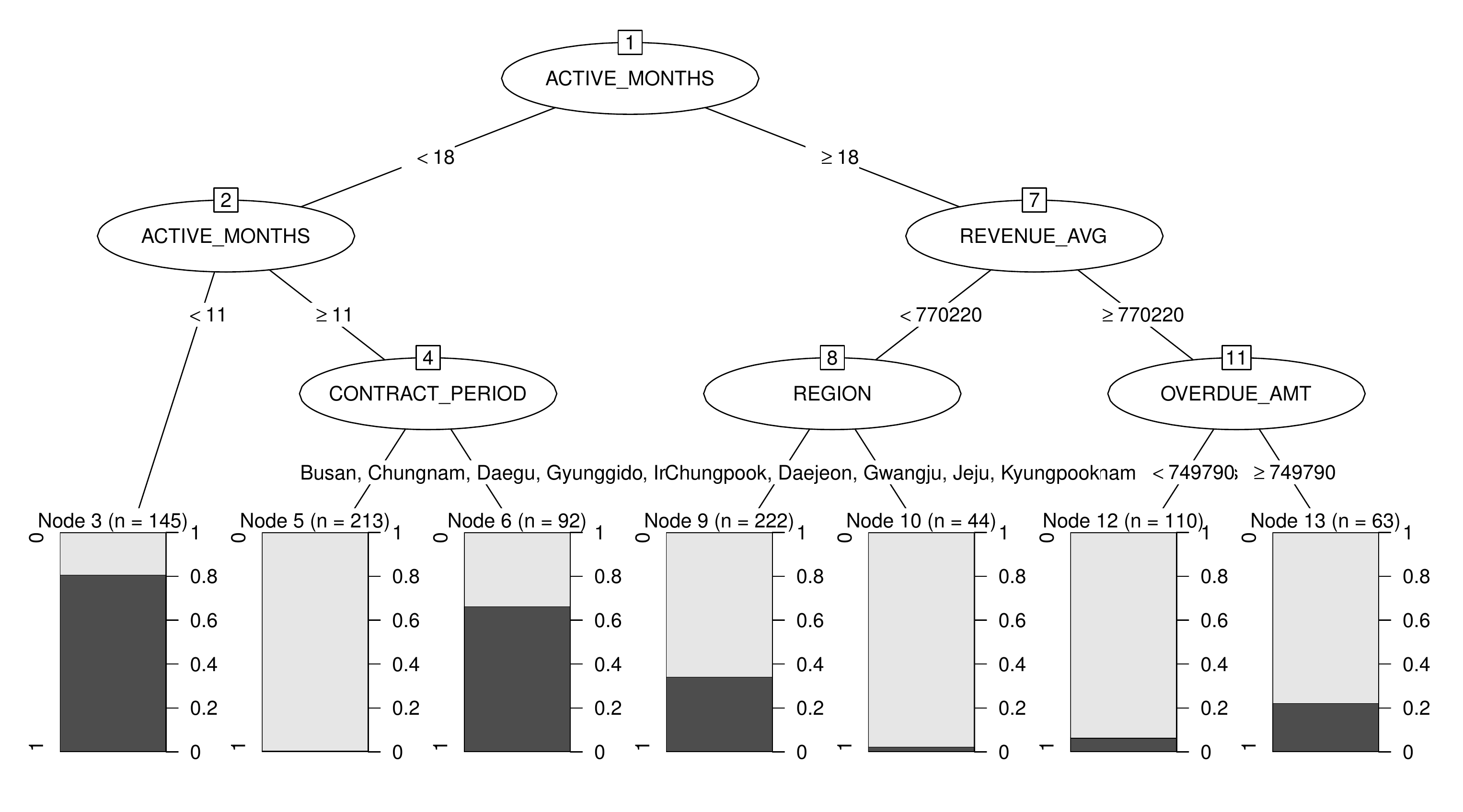}
		\caption{}
		\label{fig:ProfTree_Koreanpersonal_casestudy}
	\end{subfigure}
	\caption[Growing EvTree and ProfTree]{Trees for customer churn prediction constructed by EvTree (a) and ProfTree (b).}
	\label{fig:EvTree_ProfTree_Koreanpersonal_casestudy}
\end{figure}

EvTree's evolutionary algorithm contains a user-specified parameter $\alpha$ which plays the same role as the parameter $\lambda$ within ProfTree as it regulates the complexity of the trees that are grown. In order to find the optimal value for $\alpha$, the same method is used to find the optimal $\lambda$-value for ProfTree. Hence, we apply grid search in combination with $5\times 2$-fold cross-validation as described in Section \ref{subsec:tuninglambda}. The EMPC criterion is hereby used with its default values (as specified in Section \ref{subsec:performancemeasures}) to select the optimal value for $\alpha$.
EvTree reaches the highest average EMPC at $\alpha_{opt}=0.14$ (Figure \ref{fig:EvTree_tuningalpha_casestudy}) while ProfTree at $\lambda_{opt}=0.16$ (Figure \ref{fig:ProfTree_tuninglambda_casestudy}).
Next, we grow both EvTree and ProfTree on the entire data set with the respective optimal value of their regularization parameter (Figure \ref{fig:EvTree_ProfTree_Koreanpersonal_casestudy}).

EvTree uses the same two variables (\verb|active_months| and \verb|contract_period|) as CART and ctree to predict churners, but each method uses different cutoff values. On the other hand, ProfTree incorporates three additional variables (\verb|revenue_avg|, \verb|region| and \verb|overdue_amt|) in its classification tree. Furthermore, EvTree's first split variable is \verb|contract_period| while this variable only occurs in the third level of ProfTree's solution. When a customer stays with the telecom operator for 18 months or more, ProfTree employs variables like average revenue and overdue amount to determine the customer's likelihood to churn since losing a long-time customer may cause a bigger loss or profit than a recently joined customer.

All trees can easily be interpreted and communicated to the management professionals of the retention campaign. However, we still need to compare their performance, especially their (expected) profitability. Table \ref{tab:performances_casestudy} contains the results from the different classifiers for various performance measures.
\begin{table}[h]
	\centering
	\begin{tabular}{| l || c | c | c | c | c | c | c | }
		\hline
		& EMPC & MPC & $\overline{\eta}_p$ & $\overline{\eta}_r$ & $\overline{\eta}_F$ & AUC & MER   \\ \hline\hline
		ProfTree &  \textbf{13.769} & \textbf{13.739} & \textbf{0.520} & \textbf{0.949} & \textbf{0.672} & \textbf{0.879} & 0.178   \\ \hline
		EvTree    &  12.865 & 12.696 & 0.442 & 0.913 & 0.596 & 0.868 & \textbf{0.139}   \\ \hline 
		CART      &  12.576 & 12.407 & 0.436 & 0.899 & 0.587 & 0.855  & 0.147   \\ \hline
		ctree      &  12.584 & 12.407 & 0.436 & 0.906 & 0.589 & 0.858  & 0.147  \\ \hline
	\end{tabular}
	\caption{Performances of various tree classifiers on the Korean telecom churn data set.}
	\label{tab:performances_casestudy}
\end{table}

ProfTree has by far the highest EMPC and MPC value among all the tree classifiers. Compared to EvTree, it yields (expected) profit gains of $0.894$ EUR per customer. A telecommunication service provider has typically several thousands or even millions of customers. Hence, the potential profit gain could be enormous. Perhaps surprisingly, ProfTree even has the largest AUC. EvTree has built the tree with the lowest MER. However, as this case study shows, this does not imply that EvTree's solution is the most profitable. Furthermore, ProfTree exhibits the highest $\overline{\eta}$-precision ($52\%$), meaning it most effectively identifies churners correctly. In fact, it is the only classifier for which $\overline{\eta}_p$ is larger than $50\%$, so more than half of the $\overline{\eta}_{empc}$-based customer list contains actual would-be churners. ProfTree's solution also has the highest $\overline{\eta}$-recall ($94.9\%$), thus it is capable of detecting almost all churners.

In summary, this case study illustrates how ProfTree can be employed to build a classification tree that balances profitability and complexity by searching a larger space of potential trees. It is worth mentioning that in this setup, several runs of the ProfTree algorithm with the same parameters typically lead to the same tree. However, this may not always be the case due to the stochastic nature of the search algorithm and the potential vast search space. As a result, trees with different structures but similar fitness function values may be found by subsequent runs of ProfTree. We alleviated this problem in the case study by restricting the maximal tree depth, yielding a unique solution.

\begin{figure}[h!]
	\centering
	\includegraphics[width=0.95\textwidth]{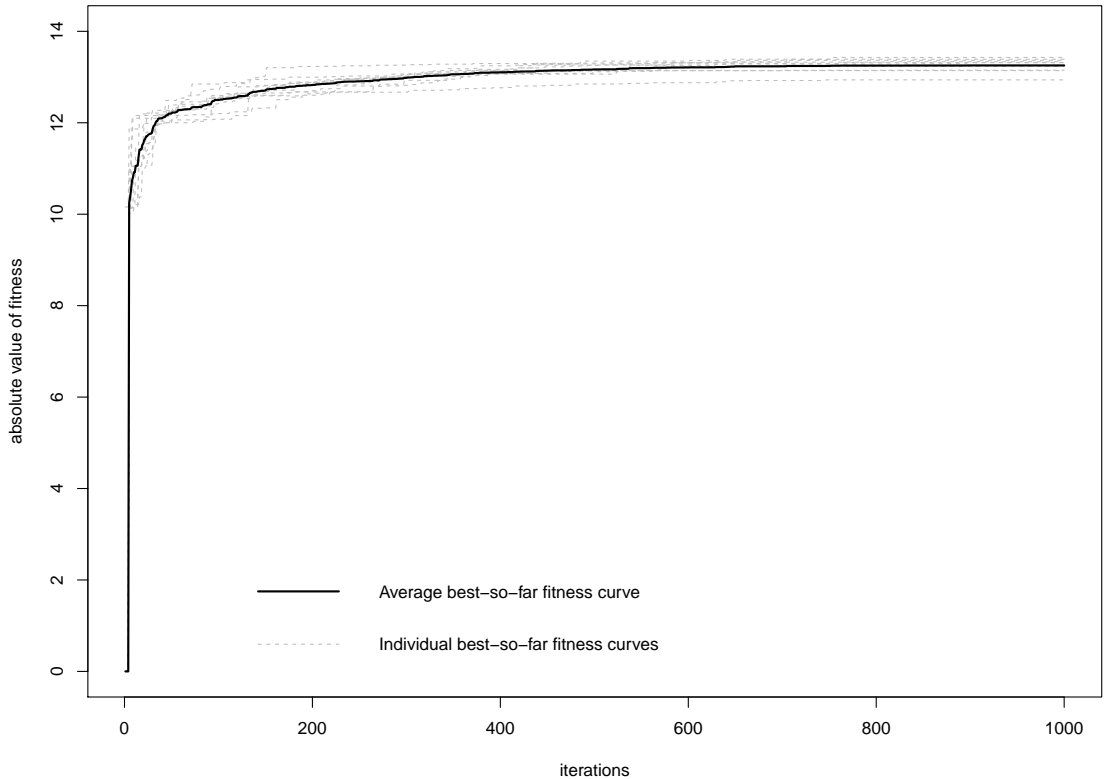}
	\caption{Fitness curve for each of 10 runs of ProfTree on the Korean churn data set.}
	\label{fig:fitnesscurves_Korean_personal_10folds}
\end{figure}

Suppose we would instead use the default settings of ProfTree: maximum tree depth of $9$, minimum of $7$ observations per terminal node and minimum of 20 observations per internal node. Figure \ref{fig:fitnesscurves_Korean_personal_10folds} shows the convergence pattern of $10$ runs of ProfTree on the Korean churn data set. Each of the $10$ fitness curves reaches the maximum solution quite fast, since the performance stabilizes already after $600$ iterations. Furthermore, each run of ProfTree leads to a different solution, but their corresponding fitness functions are very similar.

\section{Empirical evaluation\label{sec:empiricalevaluation}}

In this section, we compare ProfTree with EvTree, CART, C4.5 and ctree in a more rigorous context. Both the CART and C4.5 algorithm have the option to include a final, global pruning step to simplify the tree by snipping off the least important splits. We consider both CART and C4.5 with and without this pruning step. We apply the classification algorithms to several real-life churn data sets from various telecommunication service providers. Their performance is evaluated by using the EMPC, MPC, AUC, MER, $\overline{\eta}_p$, $\overline{\eta}_r$ and $\overline{\eta}_{F_1}$ . We begin by describing the experimental setup, followed by presenting the results of the benchmark study, and a discussion.

\subsection{Experimental setup of the benchmark study\label{subsec:simulationsetup}}

For each data set, we perform five replications of twofold cross-validation ($5\times2$ cv) \citep{dietterich1998approximate, demvsar2006statistical}. In each replication, the data set is randomly partitioned into two equalsized sets, stratified according to the churn indicator. Next, each technique is trained on each set and tested on the other set. So in total, each learning algorithm is estimated ten times on a random half of the data set and its performance is evaluated on the other half.

When preprocessing the  data, the observations (i.e. customers) with missing values are removed before the analysis. No other data preparation steps are applied because the tree-based methods do not require standardization of the continuous predictor variables or transformation of the categorical predictors. Furthermore, we do not have to remove strongly correlated features because multicollinearity among the predictors is not an issue for classification tree methods. In Table \ref{tab:summary_datasets} the main characteristics of the churn data sets are described.
\begin{table}[h!]
	\centering
	\begin{tabular}{l l l c c c}
		\textbf{ID} &  \textbf{Source} & \textbf{Region} & \textbf{\# Obs.} & \textbf{\# Att.} & \textbf{Churn rate $[\%]$}  \\ \hline\hline
		KorPer & Operator & East Asia & 889 & 10 & 31.16 \\ \hline
		Belg & Operator & Europe & 3698 & 9 & 13.28 \\ \hline
		UCI & UCI & - & 5000 & 19 & 14.14 \\ \hline
		Chile & Operator & South America & 7056 & 37 & 29.14  \\ \hline		
		Duke & Operator & North America & 12499 & 9  & 39.32  \\ \hline
		KorCor & Operator & East Asia & 13601 & 15 & 22.59 \\ \hline
	\end{tabular}
	\caption{Characteristics of churn data sets: ID, source, region, number of observations, number of attributes, and churn rate.}
	\label{tab:summary_datasets}
\end{table}

As suggested by \citet{grubinger2011evtree}, ProfTree, EvTree, CART and ctree models are constrained to a minimum number of $20$ observations per internal node, $7$ observations per terminal node, and a maximum tree depth of 9. Apart from those parameters, the default settings of the algorithms are used. C4.5 is only constrained to a minimum of 20 samples that must be put in at least two of the splits, since other restrictions are available in this implementation. Both CART and C4.5 models are considered with and without the pruning step. Furthermore, we use EMPC's default values as specified in Section \ref{subsec:performancemeasures}. For determining the optimal values for the regularization parameters $\lambda$ and $\alpha$ within ProfTree and EvTree, respectively, we apply the same technique as in the case study: grid search in combination with $5$ times $2$-fold cross-validation together with the EMPC criterion.

\subsection{Results of the experiment\label{subsec:simulationresults}}

\begin{figure}[h!]
	\centering
	\begin{subfigure}[b]{0.87\textwidth}
		\includegraphics[width=1\linewidth]{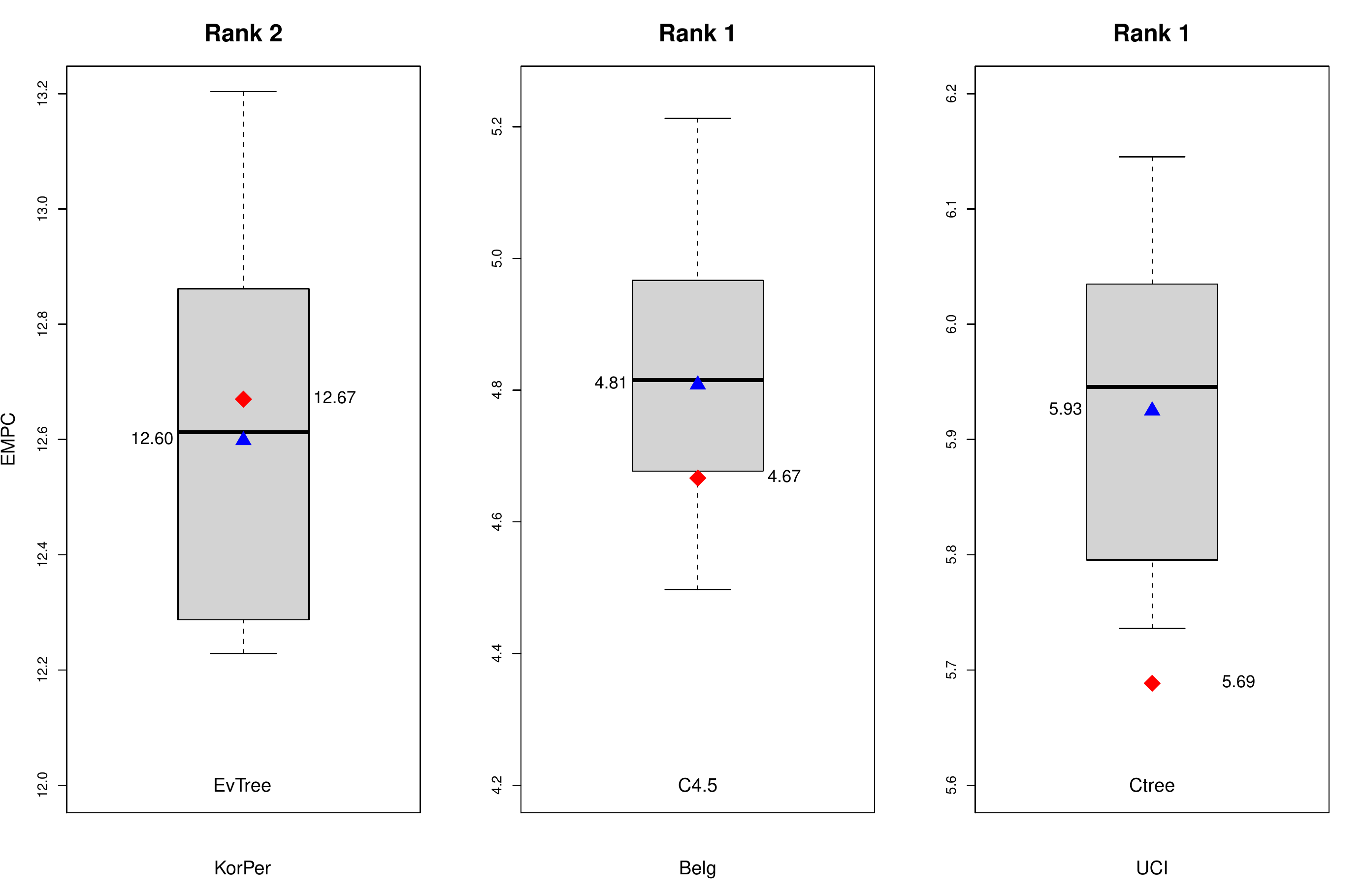}
		\caption{}
		\label{fig:5x2cv_analysis_EMPC_boxplots_v1} 
	\end{subfigure}
	\begin{subfigure}[b]{0.87\textwidth}
		\includegraphics[width=1\linewidth]{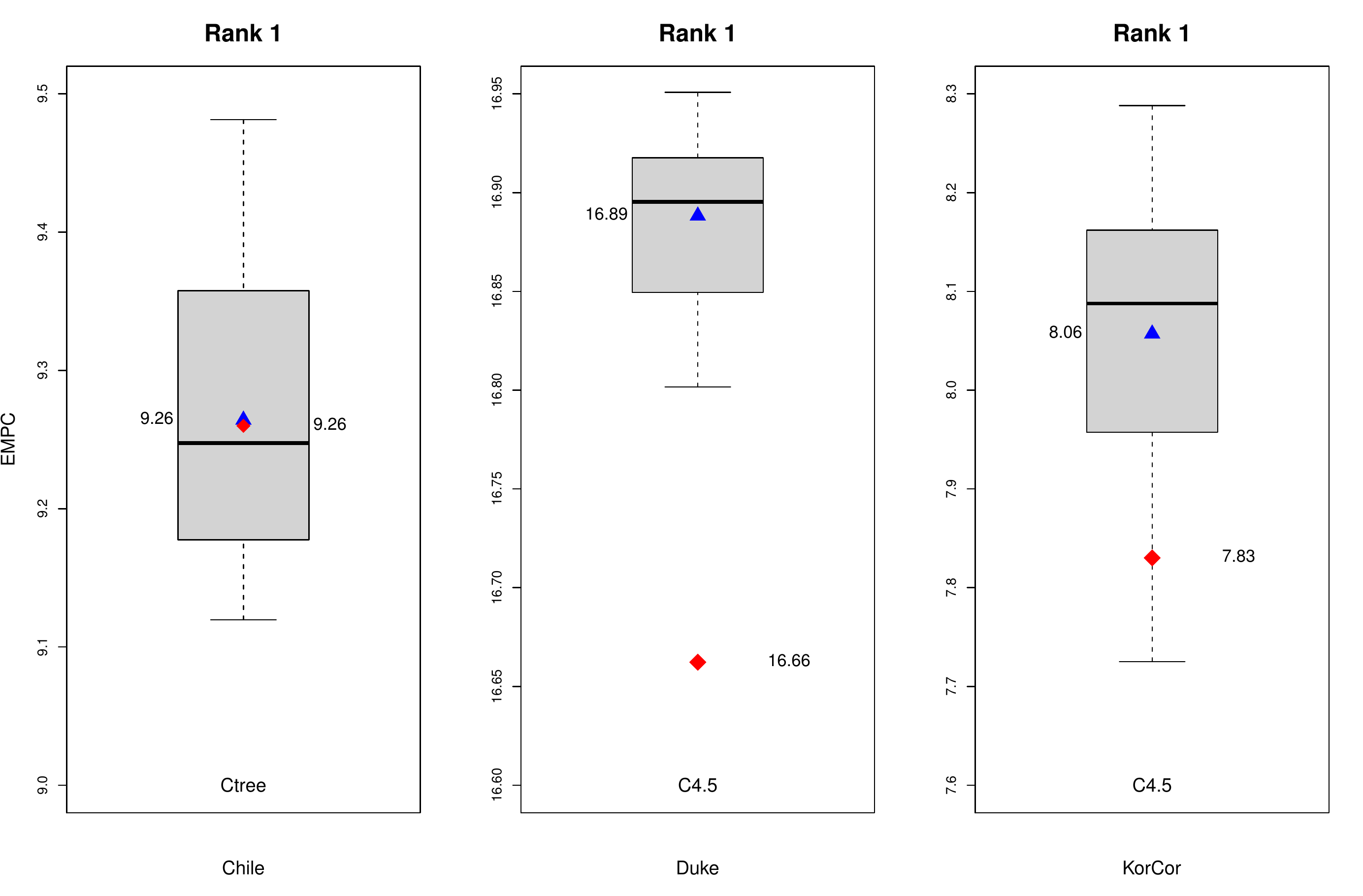}
		\caption{}
		\label{fig:5x2cv_analysis_EMPC_boxplots_v2}
	\end{subfigure}
	\caption[EMPC boxplots]{The boxplots are based on $10$ EMPC estimates of ProfTree measured on the testing folds from the $5\times2$ cross-validation. $\blacktriangle$: ProfTree; $\bigLozenge$: best competitive classifier respective to the data set (label at the bottom).}
	\label{fig:5x2cv_analysis_EMPC_boxplots}
\end{figure}

\begin{figure}[h!]
	\centering
	\includegraphics[width=0.85\textwidth]{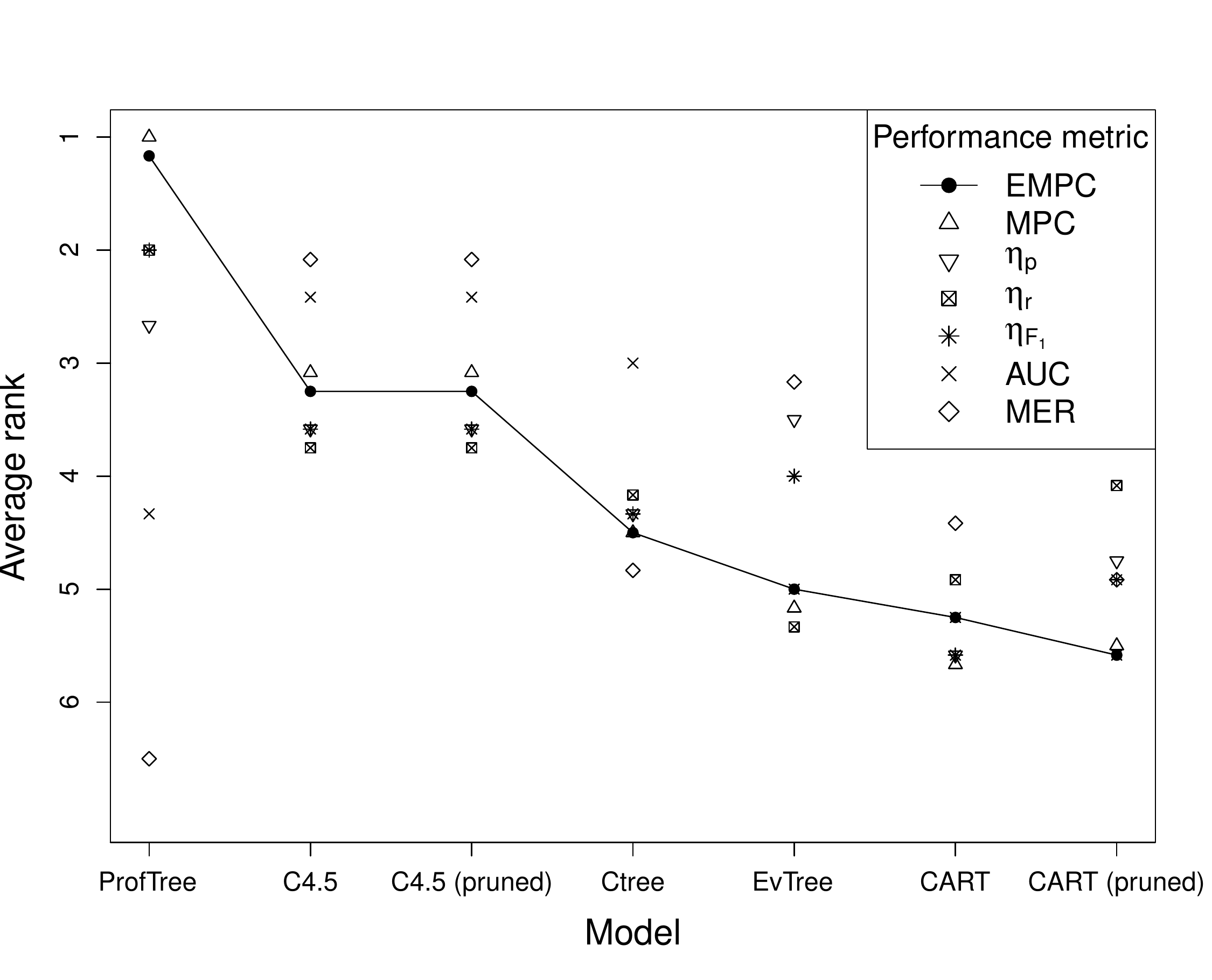}
	\caption{Average rank of the classifier over the different data sets for various performance metrics.}
	\label{fig:rank_vs_model}
\end{figure}

Figure \ref{fig:5x2cv_analysis_EMPC_boxplots} shows the boxplot of the $10$ EMPC values of ProfTree for each churn data set. We compare the average EMPC of ProfTree ($\blacktriangle$) with the one from the respective best competitive classifier ($\bigLozenge$). Note that the best competitive classifier is labeled at the bottom of each boxplot. In order to avoid overlapping labels and obscuring the view of the estimates, we only show the result from the respective best competitive technique. When averaging the ranks over the data sets, ProfTree has the overall best performance in terms of EMPC and MPC (Figure \ref{fig:rank_vs_model}). Our new method occupies the first place in five out of six data sets in terms of EMPC, resulting in an average rank of $1.17$ on a scale from $1$ to $7$. ProfTree delivers in $5$ out of $6$ data sets the most profitable churn model, and in $3$ out of the $5$ best cases the outperformance is significant. For the other data set, ProfTree falls closely behind the best competitive method. The profit gains range from 0.001 to 0.24\euro\text{ }per customer. In the case of the UCI data set, ProfTree's EMPC estimates range from 5.74 to 6.15\euro per customer, and its average performance equals $\overline{\text{EMPC}}=5.93\pm 0.14$\euro. This is on average $4\%$ better than the respective best competitive model (ctree). In the worst case (KorPer), ProfTree's EMPC estimates range from 12.23 to 13.20\euro, and its average performance equals $\overline{\text{EMPC}}= 12.60 \pm 0.32$\euro. This is on average $0.6\%$ worse than the respective best competitive classifier (EvTree). When measuring the MPC performance, ProfTree has an average rank of 1, being the best churn model in all six data sets (Figure \ref{fig:rank_vs_model}).

Furthermore, ProfTree exhibits the overall highest $\overline{\eta}$-precision (average rank: 2.67), so it most effectively identifies churners correctly. It also has the highest average $\overline{\eta}$-recall (average rank: 2.00), thus it is capable of detecing the most would-be churners. Of course, when considering the $\overline{\eta}$-based $F_1$ measure, ProfTree is also the best performing classifier with an average rank of $2.00$, being the best churn model in four out of six data sets. Evidently, ProfTree has the overall poorest MER performance, having in five out of six data sets a rank of 7 (average rank: 6.5). Because ProfTree maximizes EMPC, rather than minimizing the misclassification error, significant discrepancies between these two measures were expected. Similarly, in the case of AUC performance, ProfTree is ranked quite low with an average rank of 4.33.

\subsection{Discussion\label{subsec:discussion}}

The benchmark studies demonstrate the benefit of incorporating the EMPC, a profit-based performance measure, into an evolutionary-driven classification tree. ProfTree achieves, on average, significant profit gains compared to the competitive techniques. The study showed that ProfTree is the overall most profitable classifier in terms of the EMPC and the MPC, where, for the latter, it is the best churn model in all six data sets.

The benchmark studies further illustrate that model selection purely based on accuracy related performance measures, such as AUC and MER, likely results in considerable less profitable models. This is clearly demonstrated by the fact that ProfTree, which maximizes profit in its construction step, is the overall most profitable classifier but simultaneously has the worst MER values (see Figure \ref{fig:rank_vs_model}).

Interestingly, ProfTree also exhibits the overall highest $\overline{\eta}$-precision (see Figure \ref{fig:rank_vs_model}) despite the fact that the $\overline{\eta}$-based performance measures introduced in Section \ref{subsec:etaperformancemeasures} are related to the notion of accuracy. This is a desirable property of ProfTree because, although a company's primary objective is to maximize profits, another important requirement is that the marketers have as many true would-be churners on their target list as possible. ProfTree produces customer lists with the highest hit rates ($\overline{\eta}_p$), meaning they contain many true churners. In other words, ProfTree is the most effective classifier to correctly identify churners, which allows companies to not only focus their marketing resources on the customers that intend to churn but also to focus on those who are the most profitable to the company.

In addition to high precision, ProfTee also exhibits the overall highest $\overline{\eta}$-recall, meaning it can detect the largest proportion of would-be churners. Furthermore, all $\overline{\eta}$-based performance metrics are optimized for maximum profit, which means that churn management campaigns are able to focus their efforts primarily on customers that are profitable to the company.

\section{Conclusions and future work\label{sec:conclusion}}

In this paper, we presented a new churn classification method called ProfTree\footnote{ProfTree will become available in an R-package. In the meantime, R code implementation of ProfTree can be obtained by sending an e-mail to the corresponding author.}
 that uses an evolutionary algorithm to directly optimize the EMPC \citep{verbraken2013novel} in the model construction step of a decision tree. As a result, ProfTree aims to actively construct the most profitable model for a customer retention campaign. We exploit an evolutionary algorithm for learning profit driven decision trees according to the regularized EMPC fitness function (\ref{thetahatempc}). One major benefit of using decision trees as the underlying classifier is that the model can be easily interpreted which helps to understand why customers defect.

In our benchmark study, ProfTree is the overall most profitable model compared to $6$ other tree-based methods. The study consisted of applying the classifiers to $6$  real-life churn data sets. Five replications of $2$-fold cross-validation were used to evaluated their performance based on accuracy, cost and profit related performance metrics. We conclude that model selection based on accuracy, like MER or AUC, leads to less profitable results. In almost all cases, ProfTree outperforms its competitors, leading to significantly higher profits; whereas, in the worst case, its profit losses are relatively small compared to the respective best competitive model.

Furthermore, the benchmark study shows that ProfTree has the highest hit rate and recall rate, which makes it the most effective model to both correctly identify churners as well as to detect the largest proportion of would-be churners.
In this paper, we have shown that our proposed method aligns best with the core business requirement of profit maximization. Moreover, ProfTree produces customer lists with the highest $\overline{\eta}$-precision which can be used in the retention campaign for targeting the most profitable potential churners.

Concerning future research, we intend to combine the ProfTree algorithm with random forests \citep{breiman2001random}. This method, called ProfForest, aims at further improving the profit maximizing property by building a large collection of profit induced trees, and then aggregating them.

\section*{Acknowledgements}
This work was supported by the BNP Paribas Fortis Chair in Fraud Analytics and Internal Funds KU Leuven under Grants C16/15/068 and C24/15/001.



\section*{References}
\bibliographystyle{elsarticle-harv} 
\bibliography{References}


%
%
%
\end{document}